\documentclass[]{phai}

\usepackage{booktabs}
\usepackage{url}
\usepackage{hyperref}
\usepackage{graphicx}
\usepackage{amsmath}
\usepackage{xcolor}
\usepackage{textcomp}
\usepackage{subcaption}
\usepackage{enumitem}

\newcommand{\methodname}{TouchTwin}
\newcommand{\PAR}[1]{\vskip3pt \noindent{\bf #1~}}

\title{\methodname: Scaling Human Touch Data via Real-to-Sim-to-Real}

\author[1,2,*]{Huiwen Han}
\author[2,*]{Nuo Chen}
\author[3,*]{Ziyao Zeng}
\author[2]{Dayou Li}
\author[2]{Hao Wang}
\author[2]{Amir Hossein Rahmati}
\author[2]{Xiaoning Qian}
\author[4]{Bangya Liu}
\author[2]{Zhiwen Fan}

\affiliation[1]{Stanford University}
\affiliation[2]{Texas A\&M University}
\affiliation[3]{Yale University}
\affiliation[4]{2077AI}

\contribution[*]{Equal contribution}

\abstract{
Human demonstrations are becoming the main source of data for dexterous manipulation, and real-to-sim-to-real pipelines multiply each demonstration in simulation. These pipelines rebuild the interaction from vision, which recovers how the hand moves. Which fingers touch, when, and how hard remain unobserved, because a hand pose shows where each finger stops on the surface rather than how far it is driven to close. This gap is a force gap: on held-out recordings, a hand policy that imitates the observed pose grips with 0.31$\times$ the recorded force, and a policy trained on physics replays of vision-tracked poses grips with 1.55$\times$. \methodname{} closes the gap with measured human touch. Each selected TouchScale recording, captured with head and wrist RGB-D cameras and an 880-taxel glove, is rebuilt in MuJoCo. Coding agents write the objects as physics-tested programs, and a per-subject tactile hand replays the interaction with the recorded touch setting the contact and the closing force of each finger. In simulation, a policy trained on touch-grounded replays grips at 0.99$\times$ the recorded force and succeeds in 20/20 trials on held-out recordings and 35/36 unseen perturbations; a contact predictor trained on them misses 13\% of real contact frames, against 32\% for vision-based replays. Each replay also seeds perturbed demonstrations with simulated contact and force labels, rendered as realistic video in new views and scenes.
}

\date{\today}
\metadata[Keywords]{Real-to-sim-to-real, Dexterous manipulation, Egocentric data}
\metadata[Project Page]{\url{https://touchtwin.github.io}}

\begin{document}

\maketitle

\begin{figure*}[!htb]
  \centering
  \includegraphics[width=\textwidth]{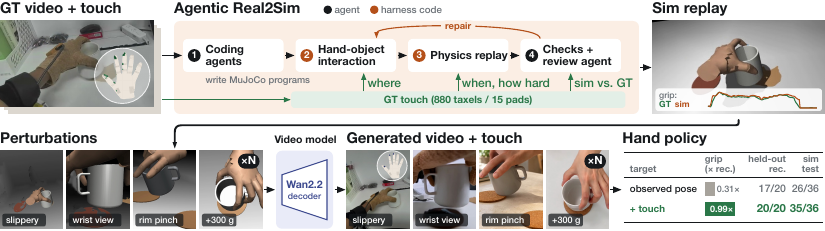}
  \caption{\textbf{\methodname{} overview.} Top: an agentic pipeline rebuilds each recording; GT touch sets where, when and how hard fingers touch and checks the replay. Bottom: each replay yields $N$ perturbed demos, decoded to video in new views and scenes, that train a hand policy.}
  \label{fig:overview}
\end{figure*}

\section{Introduction}\label{sec:introduction}

Human demonstrations are the easiest dexterous manipulation data to collect at scale, and real-to-sim-to-real pipelines rebuild them in simulation, where one recording seeds many~\citep{dexagent2026,omnihoi2026,graspandlift2026,xsim2025,videomimic2025}. However, these pipelines rebuild the interaction from vision, which recovers only how the hand moves, how the object moves, or both. Contact force never enters the simulation, since video cannot observe it~\citep{touchanything2026}. A hand pose shows where each finger stops, not how far it closes, which sets grip force, and contacts are often hidden. Video thus misses which fingers touch, when, and how hard, and its two standard uses fail in opposite directions. In our experiments, pose imitation grips with 0.31$\times$ the recorded force, and replays built from vision with 1.55$\times$. Physics alone does not recover the human grip either. In our touch-grounded replays of the mug pick-up task, the recorded grip is 1.6--3.7$\times$ the smallest force required to hold the mug. Touch measures this quantity~\citep{sundaram2019stag,touchandgo2022,opentouch2025}.

\begin{figure}[!htbp]
  \centering
  \begin{subfigure}[b]{0.238\linewidth}\centering\includegraphics[width=\linewidth]{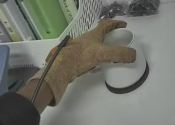}\caption{Record}\end{subfigure}\hfill
  \begin{subfigure}[b]{0.170\linewidth}\centering\includegraphics[width=\linewidth]{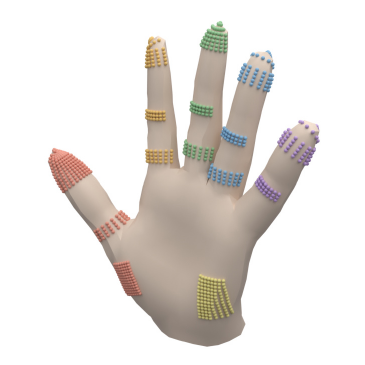}\caption{Fit}\end{subfigure}\hfill
  \begin{subfigure}[b]{0.227\linewidth}\centering\includegraphics[width=\linewidth]{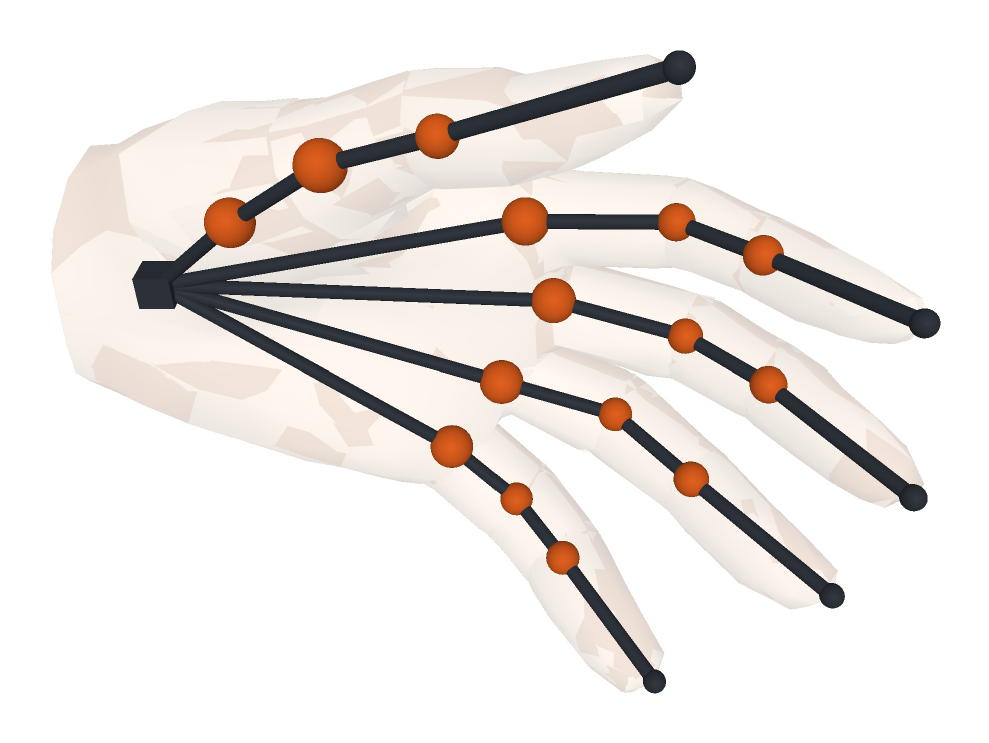}\caption{Rig}\end{subfigure}\hfill
  \begin{subfigure}[b]{0.142\linewidth}\centering\includegraphics[width=\linewidth]{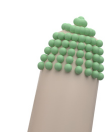}\caption{Taxels}\end{subfigure}\hfill
  \begin{subfigure}[b]{0.173\linewidth}\centering\includegraphics[width=\linewidth]{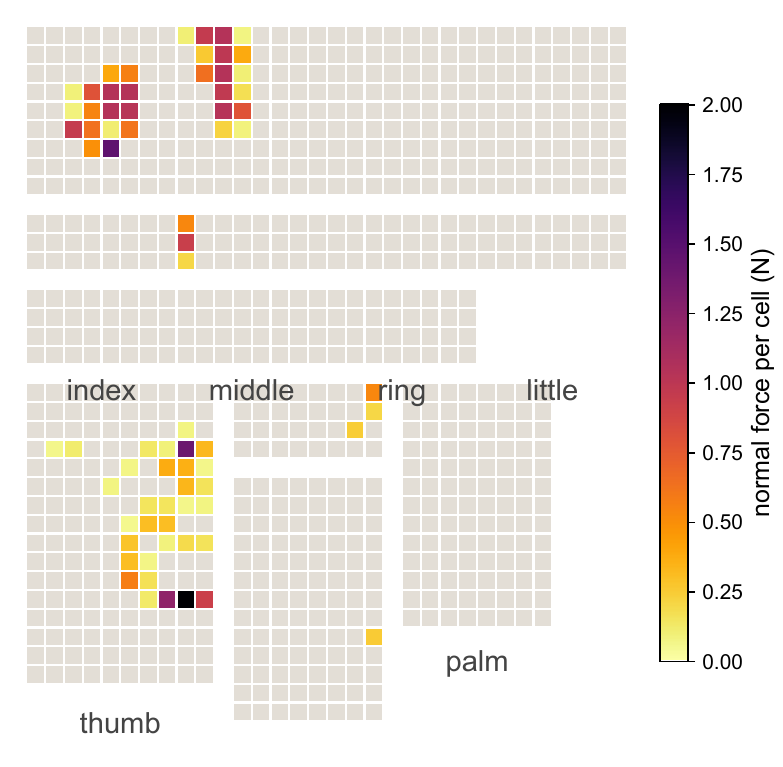}\caption{Check}\end{subfigure}
  \caption{\textbf{Simulating a tactile hand from egocentric video and paired tactile recordings.} (a)~A participant's tactile glove grasping a mug, seen by the head camera. (b)~The simulated hand fitted to the participant's episodes, with the glove's taxels on its skin. (c)~The joint chain, one dot per hinge. (d)~A fingertip pad and its taxels. (e)~The real glove's force map over its 880 cells, which the simulated touch is checked against.}
  \label{fig:hands}
\end{figure}

\begin{figure}[!htbp]
  \centering
  \captionsetup[subfigure]{justification=centering}
  \newcommand{\obj}[3]{\begin{subfigure}[t]{0.158\linewidth}\centering\includegraphics[width=\linewidth]{figures/obj_#1.pdf}\caption*{\scriptsize\textbf{#2}\\ #3}\end{subfigure}}
  \obj{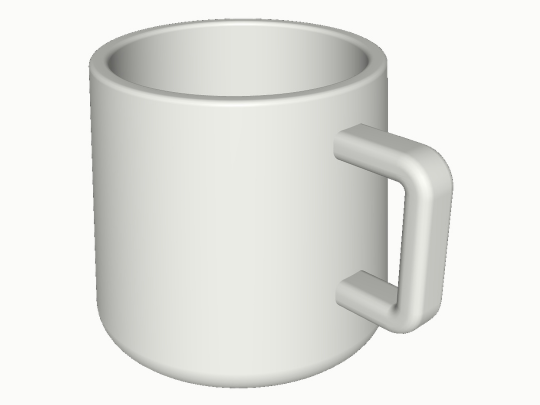}{White ceramic mug}{rigid}\hfill
  \obj{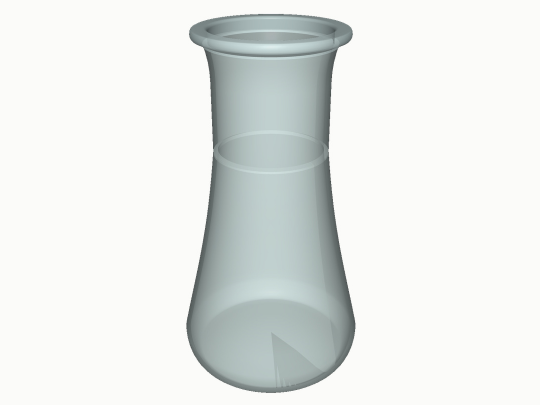}{Glass flask}{rigid, transparent}\hfill
  \obj{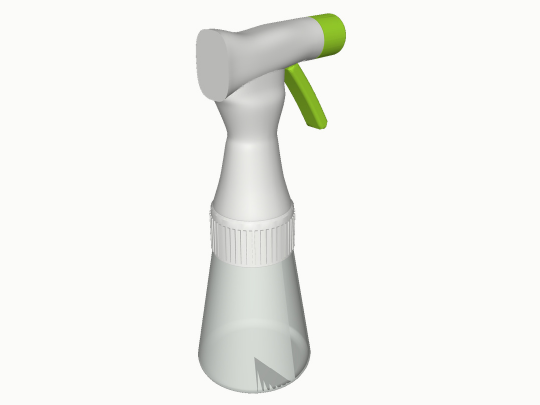}{Spray bottle}{trigger, nozzle}\hfill
  \obj{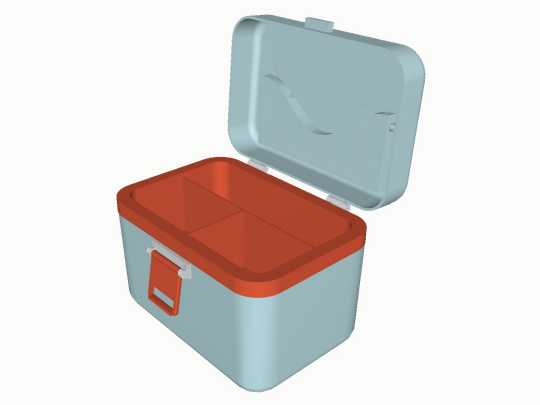}{First-aid box}{latch, lid, handle, tray}\hfill
  \obj{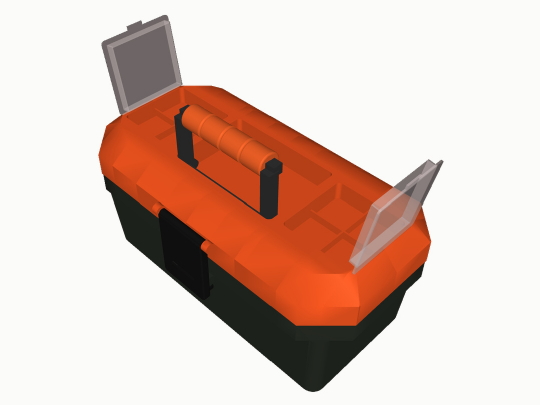}{Orange toolbox}{lid, latch, handle, 2 covers}\hfill
  \obj{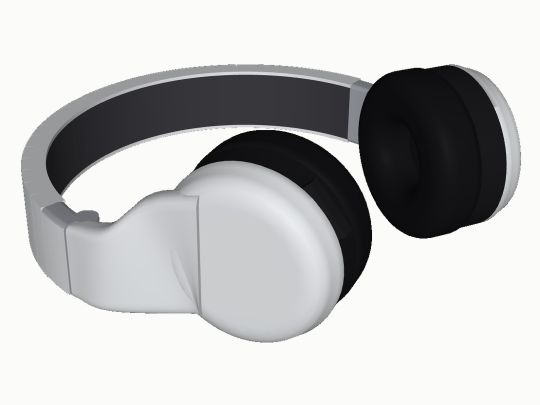}{Headphones}{fold hinges, sliders}\\[4pt]
  \obj{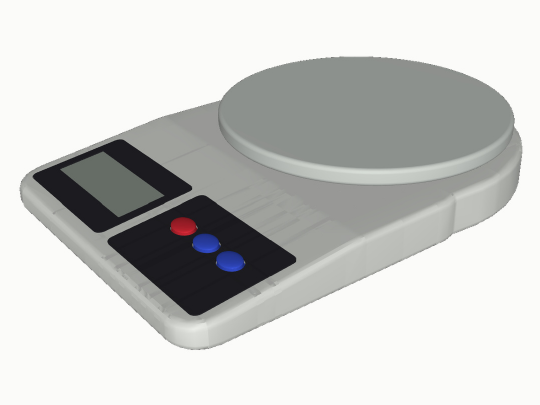}{Kitchen scale}{buttons, battery cover}\hfill
  \obj{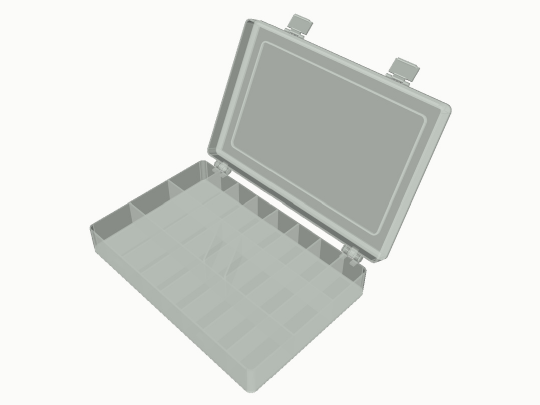}{Clear parts box}{2 latches, lid}\hfill
  \obj{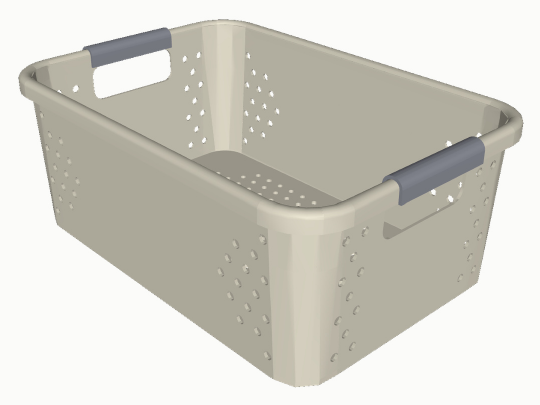}{Storage basket}{rigid}\hfill
  \obj{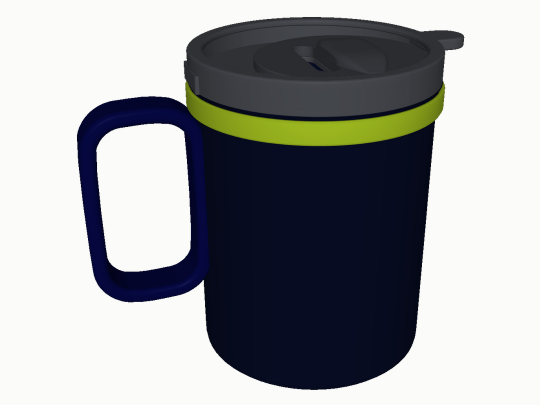}{Steel mug}{sliding drink-lid}\hfill
  \obj{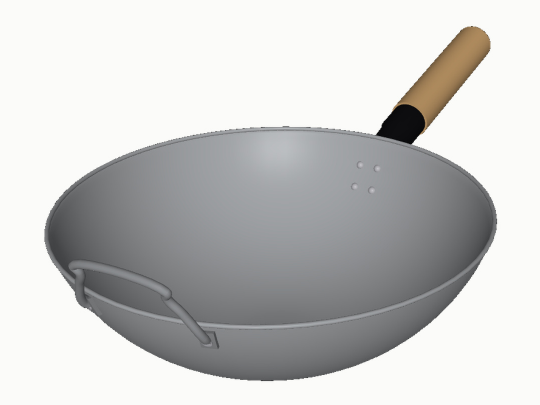}{Wok}{rigid}\hfill
  \obj{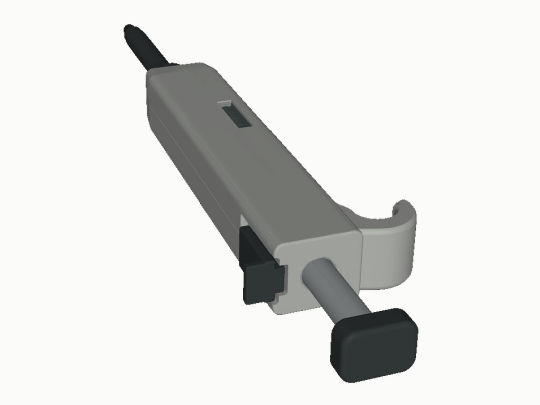}{1\,mL pipette}{plunger, tip ejector}
  \caption{\textbf{Examples of objects written by coding agents as physics-tested MuJoCo programs.} Each tile is a render of one simulation asset; the line under its name lists its moving parts. Interactive versions are on the project page.}
  \label{fig:objects}
\end{figure}

\begin{figure}[!htbp]
  \centering
  \includegraphics[width=\linewidth]{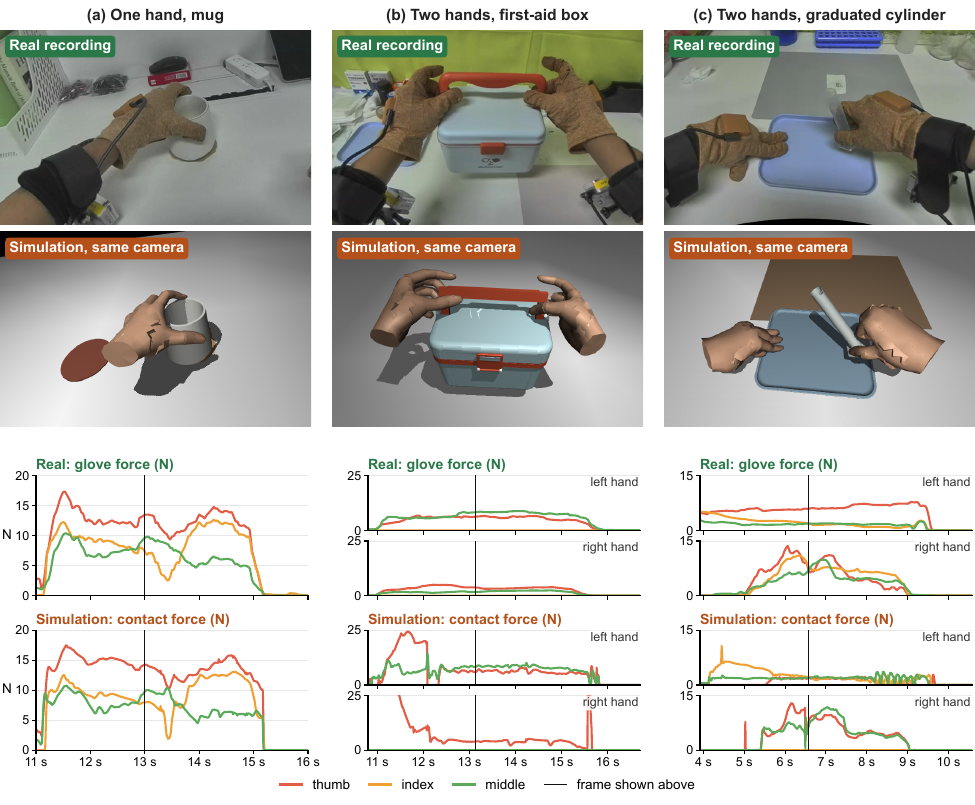}
  \caption{\textbf{Touch-grounded replays simulate the contact force.} One column per clip, real (green) above simulation (orange). Top: a frame of the real recording from the head camera, and the simulation rendered through the same camera. Bottom: the normal force of the thumb, index and middle finger over the clip, measured by the glove and simulated as contact force, on the same axes; the vertical line marks the frame shown. The object is never scripted; it moves only through contact. (a)~The left hand moves a mug to a marked spot. (b)~Both hands pinch the handle of a first-aid box, lift, carry and set it down; the simulated thumbs overshoot the glove while the grip closes and at release (peaks above 25\,N cut off). (c)~The right hand lifts a graduated cylinder out of a tray and stands it on a mat while the left hand holds the tray.}
  \label{fig:replays}
\end{figure}

Gloves and other wearables have been used to model hand-object dynamics~\citep{zhang2021dynamic}, map touch to and from video~\citep{egotactile2026,glove2hand2026,dtwm2026}, supervise robot policies~\citep{touch2robot2026,tactidex2026,mimictouch2024,dexumi2025} and identify object dynamics~\citep{forcetwin2026}, but none rebuilds the hand-object interaction as a physics simulation driven by measured touch, which lets one recording be replayed under new conditions. Touch predicted from video~\citep{touchanything2026,egotac2026,touchsight2026} is trained to reproduce glove signals from the same pixels, and it cannot recover a force the pixels do not show or change the conditions of a recording. In contrast, \methodname{} rebuilds each selected TouchScale~\citep{touchscale2026} recording in MuJoCo~\citep{todorov2012mujoco}, as shown in Figure~\ref{fig:overview}, enabling scalable human-touch data generation. Coding agents write objects as physically tested programs (Fig.~\ref{fig:objects}), a hand model per subject is tracked from both cameras~\citep{wilor2024} (Fig.~\ref{fig:hands}), and the recorded touch closes each finger past the visible pose to set where, when, and how hard it presses (Fig.~\ref{fig:replays}). In summary, we make the following contributions:
\begin{itemize}[leftmargin=1.4em,itemsep=2pt,topsep=2pt]
  \item We propose \textbf{\methodname{}}, a Real2Sim2Real pipeline that rebuilds recordings with physically tested hand-object interactions grounded by touch, and simulation carries force along with motion.
  \item We build a data engine whose replays seed perturbed demonstrations and realistic video with contact and force labels, scaling scarce touch data without additional capture.
  \item We conduct simulation experiments showing that touch fixes what vision cannot. Pose imitation grips at 0.31$\times$ the recorded force, replays built from vision at 1.55$\times$, and replays calibrated by touch at 0.99$\times$, cutting missed contact frames from 32\% to 13\%.
\end{itemize}

\section{Related Work}\label{sec:related_work}

\PAR{Real-to-sim-to-real from human demonstrations.} Recent pipelines turn human videos into simulated demonstrations for dexterous hands and humanoids: they reconstruct the hand and object from monocular or egocentric video, make the interaction physically executable, and use the simulation to train or test robot policies~\citep{dexagent2026,omnihoi2026,graspandlift2026,xsim2025,videomimic2025}. Data generation systems multiply a few demonstrations by perturbing object poses and replaying segments~\citep{mimicgen2023,dexmimicgen2025}. All of these take the interaction from vision; \methodname{} adds measured touch as the signal that decides contact and grip force.

\PAR{Human touch.} Tactile gloves record the pressure distribution of the human grasp~\citep{sundaram2019stag}, and recent datasets pair touch with vision at scale~\citep{touchandgo2022,opentouch2025,touchscale2026}. Touch has been used to model hand-object dynamics~\citep{zhang2021dynamic}, predicted from egocentric video~\citep{touchanything2026,egotac2026,touchsight2026,egotactile2026}, used to generate hand video~\citep{glove2hand2026,dtwm2026}, and used to supervise robot policies~\citep{touch2robot2026,tactidex2026,mimictouch2024,dexumi2025,n0vtla2026} and object digital twins~\citep{forcetwin2026}. These works map between sensor streams of a recording; a simulator grounded by touch can also replay the recording under new conditions.

\PAR{Simulation-conditioned video generation.} Physics engines and code have been used to drive video generation~\citep{gpt4motion2023,physgen2024,codevideomodel2026,codeworldmodel2026}, and robot-oriented video generation renders simulation as realistic video~\citep{roborender2026}. We condition Wan2.2~\citep{wan2025} on touch-grounded replays of human hands.

\section{Experiments}\label{sec:experiments}

\newcommand{\resultrow}[2]{\par\noindent\begin{minipage}[t]{0.46\linewidth}\vspace{0pt}#1\end{minipage}\hfill
  \begin{minipage}[t]{0.515\linewidth}\vspace{3pt}\centering\footnotesize\setlength{\tabcolsep}{3pt}\captionsetup{font=footnotesize,skip=4pt}#2\end{minipage}\par\vspace{6pt}}

\begin{figure}[!htbp]
  \centering
  \includegraphics[width=\linewidth]{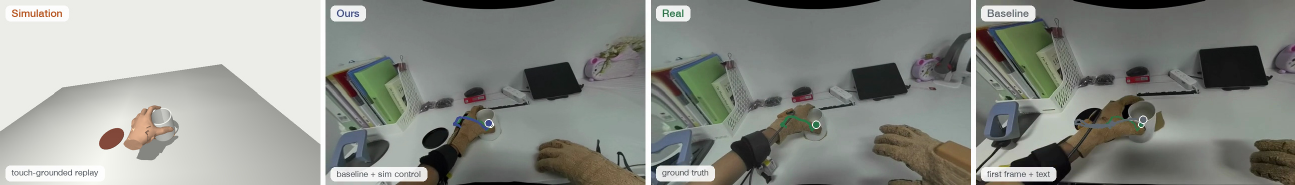}
  \caption{\textbf{Simulation-conditioned generation.} From left: the touch-grounded replay rendered through the recorded head camera, the generated video conditioned on it (baseline + sim control), the real video, and generation from the first frame and text alone, all at the same moment. Same first frame, prompt and seed; lines trace the mug, green = real.}
  \label{fig:sim2real}
\end{figure}

\resultrow{\PAR{Simulation makes generated video follow the physics.} Conditioned on the simulated hand and object, as in code- and physics-driven video generation~\citep{gpt4motion2023,physgen2024,codevideomodel2026,codeworldmodel2026,roborender2026}, Wan2.2~\citep{wan2025} generation comes closest to the real video (Table~\ref{tab:video}). Without the simulation, the generated mug takes its own path (Fig.~\ref{fig:sim2real}).}{%
  \begin{tabular}{@{}lcccc@{}}
    \toprule
    video generation & PSNR$\uparrow$ & SSIM$\uparrow$ & LPIPS$\downarrow$ & DINO$\uparrow$ \\
    \midrule
    sim render only                 & 7.79  & 0.309 & 0.757 & 0.537 \\
    first frame + text (baseline)   & 15.36 & 0.663 & 0.489 & 0.854 \\
    baseline + depth control        & 18.55 & 0.759 & 0.276 & 0.914 \\
    \textbf{baseline + sim control} & \textbf{18.89} & \textbf{0.766} & \textbf{0.258} & \textbf{0.928} \\
    \bottomrule
  \end{tabular}
  \captionof{table}{\textbf{Generated vs.\ real video} (same first frame, prompt and seed).}\label{tab:video}}

\resultrow{\PAR{Touch-grounding is essential for contact accuracy.} A contact predictor trained on vision-based simulations misses 32\% of the real contact frames on held-out recordings, whereas training on touch-grounded simulations reduces miss rate to 13\% (Table~\ref{tab:contact}).}{%
  \begin{tabular}{@{}lccc@{}}
    \toprule
    contact prediction, & contact & missed & press dur. \\
    trained on & F1$\uparrow$ & contact$\downarrow$ & err.$\downarrow$ \\
    \midrule
    vision-based replays                   & 0.778 & 32.2\% & 1.01\,s \\
    \textbf{touch-grounded replays (ours)} & \textbf{0.890} & \textbf{13.1\%} & \textbf{0.61\,s} \\
    \bottomrule
  \end{tabular}
  \captionof{table}{\textbf{Contact prediction on held-out human-ego recordings.} Press dur.\ err.: $|$predicted $-$ true$|$ press time per finger.}\label{tab:contact}}

\resultrow{\PAR{Touch supplies information beyond surface-level hand positions.} A hand policy (N$_0$-VTLA~\citep{n0vtla2026}) trained on the observed finger pose grips held-out recordings with 0.31$\times$ the true force (Table~\ref{tab:policy}); the touch-grounded replays restore human-level grip (0.99$\times$, 20/20).}{%
  \begin{tabular}{@{}lccc@{}}
    \toprule
    hand policy, & held-out & grip vs. & simulated \\
    trained on & recordings$\uparrow$ & recording & test$\uparrow$ \\
    \midrule
    observed finger pose                   & 17/20 & 0.31 & 26/36 \\
    vision-based replays                   & 15/20 & 1.55 & 33/36 \\
    \textbf{touch-grounded replays (ours)} & \textbf{20/20} & \textbf{0.99} & \textbf{35/36} \\
    \bottomrule
  \end{tabular}
  \captionof{table}{\textbf{Hand policy on held-out recordings and simulated tests}; success: slip $\le$5\,mm, set down $\le$30\,mm off.}\label{tab:policy}}

\FloatBarrier
\section{Downstream Applications}\label{sec:applications}

Each replay also seeds perturbed demonstrations with simulated contact and force labels, rendered as realistic video in new views and scenes.

\begin{figure}[!htbp]
  \centering
  \includegraphics[width=\linewidth]{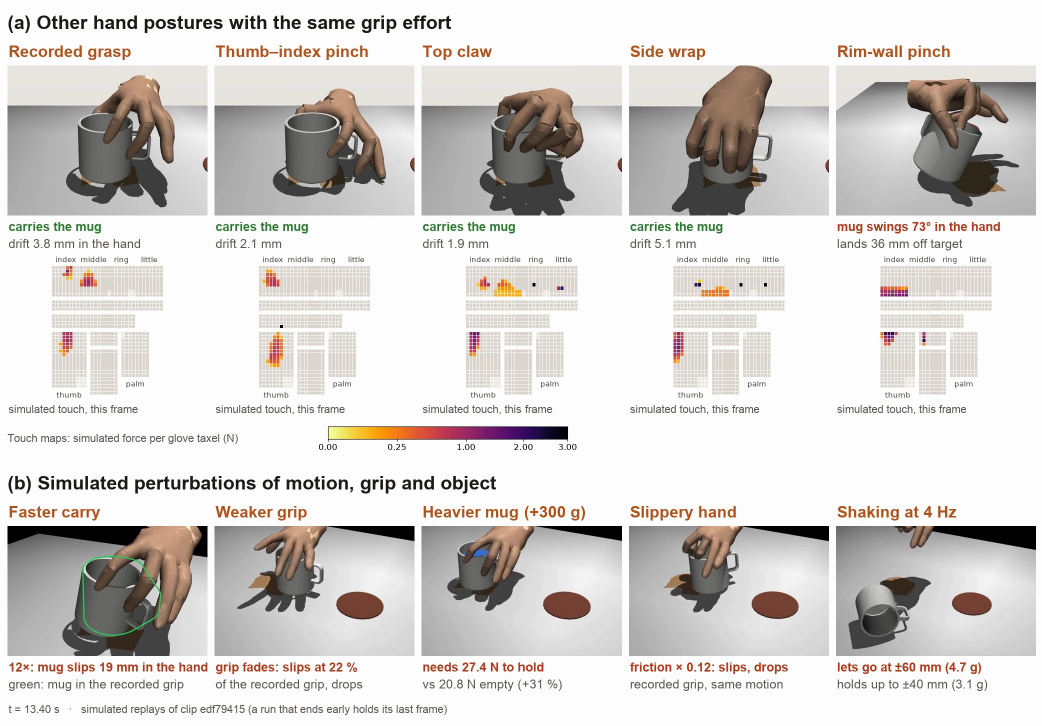}
  \caption{\textbf{Perturbed replays of one recording.} (a)~Other hand postures with the same grip effort; under each, the simulated touch of that frame on the glove's pad layout. The recorded grasp, thumb-index pinch, top claw and side wrap carry the mug (drift 3.8, 2.1, 1.9 and 5.1\,mm in the hand); with the rim-wall pinch the mug swings 73\textdegree{} in the hand and lands 36\,mm off target. (b)~Perturbations of motion, grip and object. A 12$\times$ faster carry slips the mug 19\,mm in the hand (the green outline is the mug in the recorded grip); the grip fades until the mug slips at 22\% of the recorded grip and drops; with 300\,g of water the mug needs 27.4\,N to hold against 20.8\,N empty (+31\%); at 0.12$\times$ friction it slips and drops; a 4\,Hz shake makes the hand let go at $\pm$60\,mm (4.7\,g).}
  \label{fig:perturb}
\end{figure}

\begin{figure}[!htbp]
  \centering
  \begin{subfigure}[b]{0.245\linewidth}\includegraphics[width=\linewidth]{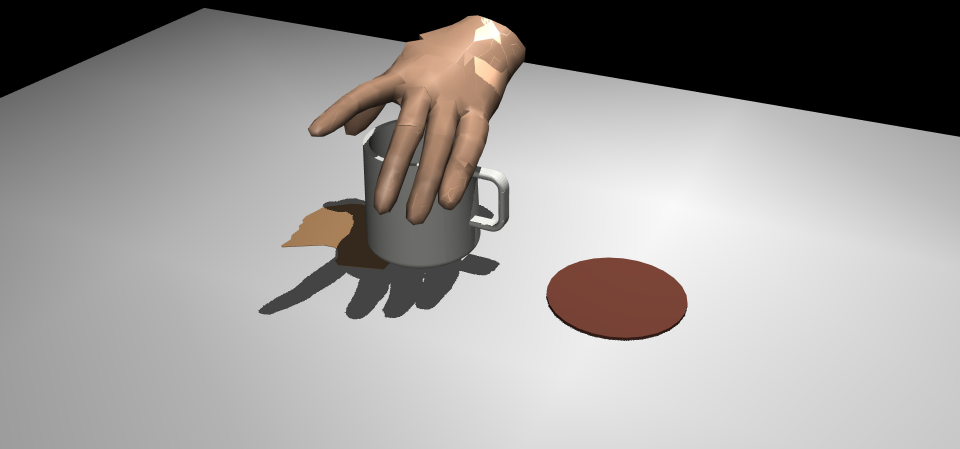}\caption{Grip fades}\end{subfigure}\hfill
  \begin{subfigure}[b]{0.245\linewidth}\includegraphics[width=\linewidth]{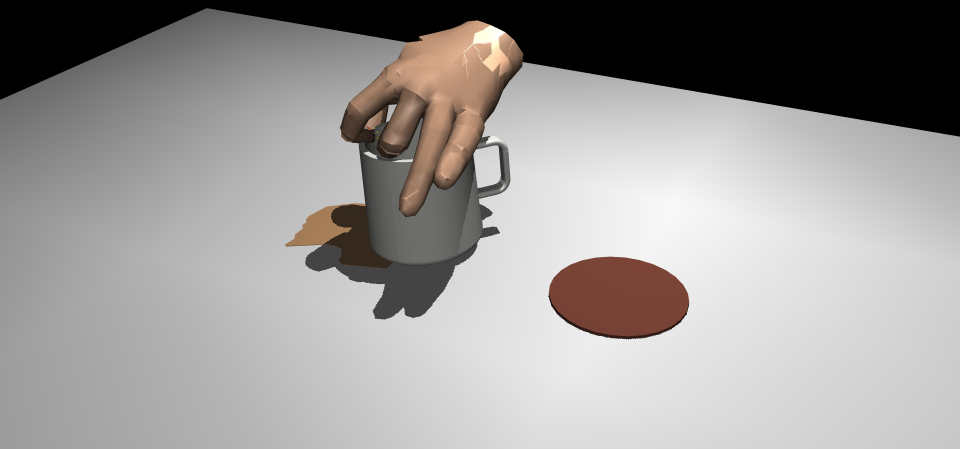}\caption{Hand gets slippery}\end{subfigure}\hfill
  \begin{subfigure}[b]{0.245\linewidth}\includegraphics[width=\linewidth]{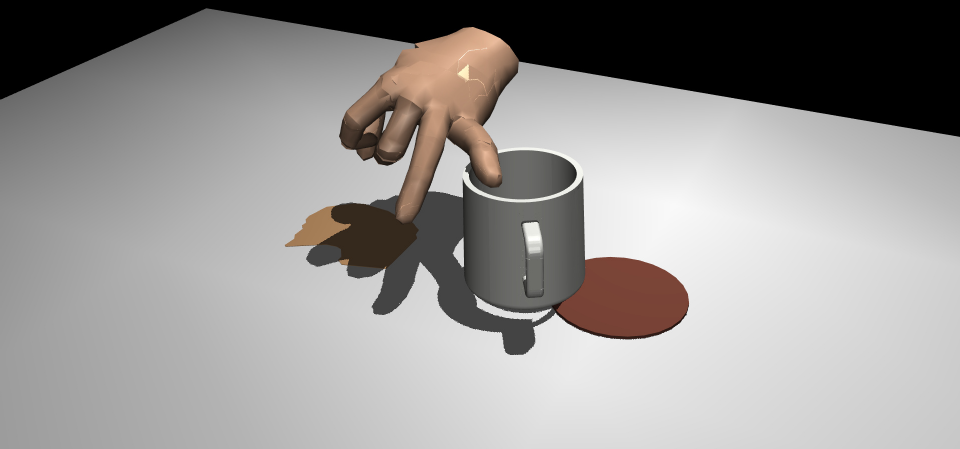}\caption{Mug gets heavier}\end{subfigure}\hfill
  \begin{subfigure}[b]{0.245\linewidth}\includegraphics[width=\linewidth]{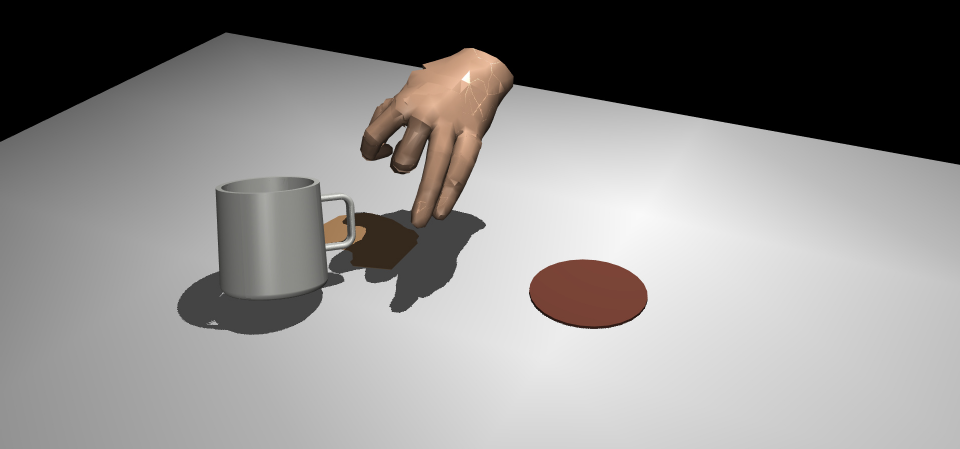}\caption{Shake builds up}\end{subfigure}
  \caption{\textbf{Perturbed until the mug drops.} During the carry, (a)~every finger's force target is turned down from 1$\times$ to 0, (b)~the hand-mug friction is turned down, (c)~mass is added up to 7.5\,kg (far more than water could add), and (d)~a 4\,Hz side-to-side shake grows in amplitude. Each sweep runs until the mug leaves the hand.}
  \label{fig:drops}
\end{figure}

\PAR{Physics turns one recording into labeled successes and failures.} A perturbation changes one thing in a replay and re-runs the physics: the grasp (another hand posture at the same grip effort), the motion (a faster carry or an added shake), the grip (each finger's force target scaled down) or the object (heavier, or more slippery). With nothing changed, it reproduces the replay exactly. Because physics resolves the outcome, the mug can slip, swing or drop, and every perturbed replay carries simulated per-taxel touch, contact and force in the glove's format (Fig.~\ref{fig:perturb}). Sweeping one factor until the object drops measures how much margin the recorded grasp had (Fig.~\ref{fig:drops}).

\begin{figure}[!htbp]
  \centering
  \includegraphics[width=\linewidth]{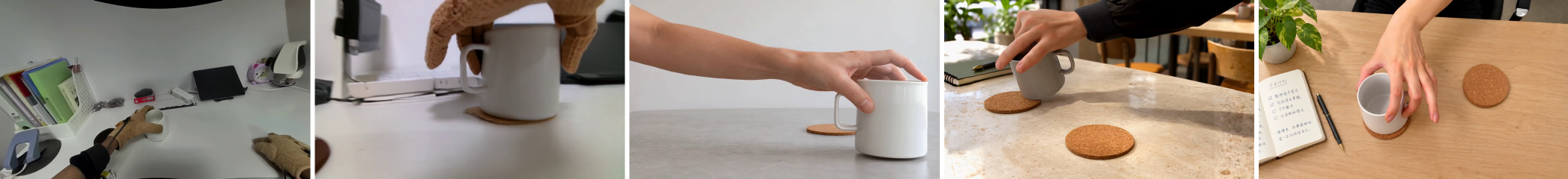}
  \caption{\textbf{Generated videos decoded from touch-grounded replays} (one frame each). From left: a generated video in the recording's own view and scene; a wrist camera in the same scene; new viewpoints, hand postures and scenes.}
  \label{fig:generated}
\end{figure}

\PAR{One replay becomes realistic video in new views and scenes.} Wan2.2~\citep{wan2025} renders a replay or perturbed replay as realistic video, conditioned on the simulation along the recorded head-camera trajectory. Changing the camera, the hand posture or the scene gives the same simulated interaction from other viewpoints, with other grasps and in other places, and each generated video inherits the contact, touch and force labels of its replay (Fig.~\ref{fig:generated}).

\begin{figure}[!htbp]
  \centering
  \includegraphics[width=\linewidth]{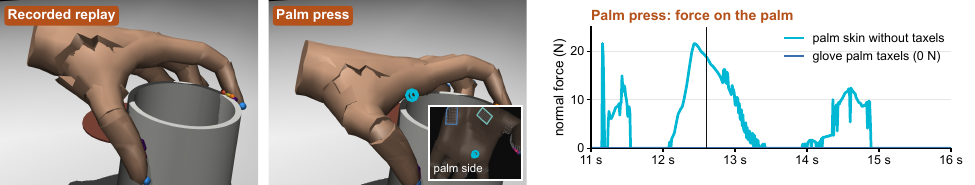}
  \caption{\textbf{Touch beyond the glove.} Left: the recorded replay, where the palm does not touch the mug. Middle: a simulated palm press, with the wrist moved 8\,mm along the palm normal past the first visible palm contact and the fingers kept on their spots; inset: the palm side, with the glove's palm pads outlined. Cyan rings: contact on palm skin the glove does not cover. Right: during the press, the palm skin carries up to 22\,N while the glove's palm taxels read zero; the vertical line marks the frame shown.}
  \label{fig:palm}
\end{figure}

\PAR{Simulation measures touch where the glove has no sensors.} The glove does not cover the whole hand: the centre of the palm and parts of the finger bases have no sensors. The simulated hand has skin everywhere, so a replay records contact that the glove cannot. Figure~\ref{fig:palm} perturbs the mug replay so that the palm presses onto the mug: the glove-layout palm taxels stay at zero, while the simulator reports the normal force on palm skin without taxels.

\FloatBarrier
\section{Limitations}\label{sec:limitations}

All policies act on simulated assets and are tested open-loop in simulation; transfer to robot hands, real-robot experiments and task diversification are still needed. The held-out sets are small, the policy and contact experiments use the mug task only, and the held-out clips start with the hand already in contact, so contact onset is not tested. Two-hand replays still slip 12--23\,mm in the hand. Generated videos inherit the simulation's errors, and the video model can still change details the simulation does not control.

\section*{Acknowledgments}
This work used the VISION supercomputer provided by Texas A\&M High Performance Research Computing (HPRC). We thank AMD for providing access to AMD Instinct MI300X GPUs through the AMD University Program and the AMD Developer Cloud.

\FloatBarrier
\bibliographystyle{assets/plainnat}
\bibliography{paper}

\end{document}